\documentclass[lettersize, conference]{ieeeconf}  

\IEEEoverridecommandlockouts                              

\usepackage{graphics} 
\usepackage{epsfig} 
\usepackage{mathptmx} 
\usepackage{times} 
\usepackage{amsmath} 
\usepackage{amssymb}  
\usepackage{mathtools}
\usepackage{bm}
\usepackage{xcolor} 
\usepackage{soul}
\usepackage{subcaption}
\usepackage[utf8]{inputenc}
\usepackage{svg}
\usepackage{balance}
\usepackage{cite}

\usepackage{blindtext}
\usepackage{csquotes}
\usepackage[normalem]{ulem}
\usepackage{comment}

\usepackage{graphicx}
\graphicspath{{./Immagini/}}

\title{\LARGE \bf
    Optimal Prioritized Dissipation and Closed-Form Damping Limitation Under Actuator Constraints for Haptic Interfaces
}

\author{Camilla Celli$^{1,2}$, Andrea Bini$^{1,2}$, Valerio Novelli$^{1,2}$, Alessandro Filippeschi$^{1,2}$,\\ Francesco Porcini$^{1,2}$ and Antonio Frisoli$^{1,2}$
\thanks{*Authors have equal contributions}
\thanks{$^{1}$Scuola Superiore Sant'Anna, IIM Institute, PERCRO Laboratory, Pisa, Italy}
\thanks{
$^{2}$Scuola Superiore Sant’Anna, Department of Excellence in Robotics and
AI, Pisa, Italy
        {\tt\small francesco.porcini@santannapisa.it}}%
}

\begin{document}

\maketitle
\thispagestyle{empty}
\pagestyle{empty}


\begin{abstract}

In haptics, guaranteeing stability is essential to ensure safe interaction with remote or virtual environments. One of the most relevant method at the state-of-the-art is the Time Domain Passivity Approach (TDPA). However, its high conservatism leads to a significant degradation of transparency. Moreover, the stabilizing action may conflict with the devices’ physical limitations. State-of-the-art solutions have attempted to address these actuator limits, but they still fail to account simultaneously for the power limits of each actuator, while maximizing transparency. 
This work proposes a new damping limitation method based on prioritized dissipation actions. It priorities an optimal dissipation direction that minimizes actuator load, while any excess dissipation is allocated to the orthogonal hyperplane. 
The solution provides a closed-form formulation and is robust in multi-DoF scenarios, even in the presence of actuator and motion anisotropies.
The method is experimentally validated using a parallel haptic interface interacting with a virtual environment and tested under different operating conditions.
\end{abstract}


\section{INTRODUCTION}

Haptics represents a growing research field both in the remote environment interaction (i.e. teleoperation) and in the virtual environment interaction. In those applications, the presence of a closed-loop force feedback aims to provide the user with the interaction information. However, this leads to well-known stability issues due to phenomena like sampling, quantization and active virtual environments or time delay \cite{diolaiti2006stability}. To provide the user with a safe interaction, stability must be guaranteed. In literature, it is often preferred to guarantee the passivity of the system (which is a more stringent condition) rather than stability, since its easier energy-based verification criteria. Thus, many passivity-based approaches have been developed over the years that aim to guarantee the stability of haptic interfaces and teleoperation architectures \cite{hannaford1989design, anderson1988bilateral, niemeyer1991stable, lawrence1992stability, hokayem2006bilateral, willaert2009bounded}.

One of the most diffused at the state-of-the-art is the \textit{Time Domain Passivity Approach} (TDPA) \cite{hannaford2002time,ryu2004sampled, hertkorn2010time, ryu2007stable}. This approach relies on two elements, \textit{Passivity Observers} (PO) and \textit{Passivity Controllers} (PC), to monitor the energy of potentially active blocks and dissipate it if active behaviour is detected, thus guaranteeing passivity. The simplicity, the wide applicability and the model-free nature of this approach make it one of the most used in literature at the cost of extreme conservatism (common in all passivity-based approaches). This results in a prominent degradation of transparency while privileging stability. In fact, transparency and stability are well-known competitive targets \cite{lawrence1992stability}.

In human-robot interaction applications, stability must be preferred over transparency for safety reasons. To ensure passivity, the PC acts as a high-frequency virtual damper that dissipates the observed energy, modifying the control reference signal (force for impedance-based devices like haptic interfaces). This stabilizing action attempt to perform the dissipation in one sample time, often clashing with the physical capabilities of the devices. In fact, power-limited actuators prevent from dissipating any observed amount of energy in one sample. Thus, a maximum amount of dissipable energy should be determined each sample to keep trace of the undissipated energy, which is collected to be dissipated in the next samples \cite{hannaford2002time}. This problem is well acknowledged at the state-of-the-art \cite{ott2011subspace}, and is potentially harmful for the architecture: dissipating actions required by the PC that exceed power-limits cannot be realized by the actuators and the stability cannot be guaranteed.

Thus, it is fundamental to define the energy that a PC can dissipate in one sample according to the physical capabilities of the system (i.e. power limits of the actuators and robot configuration). This problem can be reported to limit the damping applied by the PC. In the literature, the damping applied by the PC has been often limited according to the time-delay \cite{hertkorn2010time} (with consequent identification-related challenges, in particular in case of variable time delay) and to sample time \cite{ott2011subspace}. Time-based limitations have been shown to be ineffective, easily falling into issues \cite{porcini2023actuator}. In fact, a system constrained by high actuator power limits and short sampling time dissipates less energy than it potentially could. Conversely, when actuator power limits are low and sampling time is long, the system attempts to dissipate more energy than what can actually be achieved.

Other works face the same problem by limiting directly the rendered force to avoid violation of the passivity conditions \cite{lee2009adjusting,kim2014force}. These approaches, known respectively as Adjusting Output Limiter (AOL) and Force Bounding Approach (FBA), define the maximum applicable rendered force (thus a maximum damping to reduce the reference force) on the basis of the properties of the haptic interface. Even if the configuration of the robots is taken into account, AOL and FBA are severely model-based, requiring the knowledge of phoenomena that change in time (like friction and viscous terms) and are affected by uncertainties (like inertia). Moreover, none of those approaches take into account the power limits of the actuators, with consequent potential issues: in fact, a limited rendered force to not violate the passivity condition not necessarily doesn't violate actuators' constraints.

Thus, a promising approach seems to be considering both the robot configuration and the power limits of the actuators. In \cite{ott2011subspace}, Ott et al. imposed an inequality constraint that takes also the power limits into account, but it requires the online solution of a quadratic programming (QP) problem, requiring bulk computation. More recently, it was proposed a limitation of the damping that aimed to consider both the physical limits of the actuators and the robot configuration, while providing a closed-form solution \cite{porcini2022optimal, porcini2023actuator}. This solution has been shown very promising, however, the formulation suffers from issues related to the norm-based scalar damping limitation. In fact, in multi-DoFs systems, the method is effective when actual velocities (which are the inputs to the impedance-based PC) and actuators' limits are isotropic, i.e. when each motor has similar usage while rendering and similar power limits. Conversely, this strategy fails in presence of anisotropic input velocities (e.g. moving along only one direction) or different actuators' power limits (i.e. different joint sizes). In fact, the scalar-limited damping ensures that the norm of the dissipation vector never exceeds the norm of the maximum applicable force by the actuators, but it doesn't give any warranty on the individual component in presence of different joint sizes. Similarly, an anisotropic velocity demands dissipation primarily along the direction of motion; however, constraining this dissipation by the norm of the power limits vector (which includes the power limits over all directions) inappropriately increases the limit over the actual motion axis. It is then straightforward that to take into account multi-DoFs limits requires the choice of a dissipation direction.


The objective of the present work is to go further with respect to the state-of-the-art methods defining a new damping limitation strategy based on an optimal dissipation direction. As an optimality criterion adopted to identify this direction, there is the load on the actuators. Thus, a strategy is proposed that prioritizes the dissipation in a minimum-actuator-load direction and that defines a maximum energy dissipable in that direction. Any excess dissipation demand is then redirected to the orthogonal hyperplane (without interfering with the primary direction) up to its own defined maximum capacity. Limiting the energy dissipable in the prioritized direction and limiting the remaining in the orthogonal hyperplane means defining two maximum dampings applicable by the PC. These two limitations defined in accordance with this strategy are closed-form (no online optimization is required), and take into account power-limits over all axes of the system and are robust to the variability of the PC input vector (i.e. they are suitable for multi-DoFs devices without anisotropy-related limitations). In this work, the optimal dissipation direction to relieve the actuators is the rendering direction: the dissipating action, if directed oppositely to the rendered vector, ``erases" the reference reducing the demand on the actuators. This direction was already identified by Preusche et al. \cite{preusche2003time} as enhancing transparency. However, the present formulation is conceptually different (aiming to stability rather than transparency), offering a more comprehensive and technically precise formulation that incorporates the limitation of the PC not taken into account by \cite{preusche2003time}.


The paper is organized as follows: Section \ref{sec:mat} provides an overview of the most relevant state-of-the-art works, highlights their limitations, and presents the proposed dissipation method along with the corresponding limitation. Section \ref{sec:exp} describes the experimental setup and the experiments conducted to demonstrate the effectiveness of the method and the advances over the state-of-the-art methods. Finally, the paper concludes with a discussion and conclusions in Section \ref{sec:disc}.

\section{MATERIAL AND METHODS} \label{sec:mat}

In this section, the TDPA framework for stabilizing haptic interaction is briefly recalled, with special emphasis on the most recent damping limitation strategies. Also, the issues with the current damping limitation are discussed and supported with the evidence of ad hoc experiments to show their concrete effect. Finally, the new priority-based damping limitation strategy is presented.

\subsection{Time-Domain Passivity Approach for Haptic Interfaces}

The \textit{TDPA} is designed to enforce passivity on those architectural components that may display active behaviour, thereby ensuring stability. The complete system is modelled as an interconnected 2-port network, allowing the energy exchange across blocks to be tracked via \textit{Passivity Observers (PO)} and regulated through \textit{Passivity Controllers (PC)}. Without loss of generality, the system can be schematized as composed by a human operator controlling an impedance-based haptic interface interacting with a virtual wall. 

As shown in Figure \ref{fig:schemaablocchi}, let $\mathbf{v}(k)$ denote the velocity vector of the haptic interface, while $\mathbf{\hat{f}}(k)$ represents the force vector generated by the virtual environment. The force–velocity pairs are defined such that the power entering each block remains positive. The \textit{PO} is then expressed as follows:

\begin{figure}[t]
    \centering
    \includegraphics[width=0.45 \textwidth] {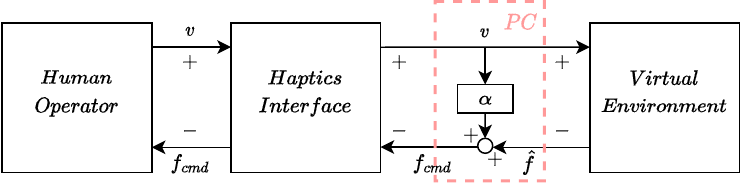}
    \caption{Block representation of an impedance-based haptic interface interacting with a virtual environment. The human operator is supposed to be passive, while the virtual environment is simulated to behave actively according to \cite{ryu2004sampled}. The PC is included to make the entire system passive.}
    \label{fig:schemaablocchi}
\end{figure}

\begin{equation}\label{eq:Ws}
E_{obs}(k) = E_{obs}(k-1) + T_s \mathbf{\hat{f}}^{T}(k)\mathbf{v}(k) + E_{PC}(k-1)
\end{equation}

$E_{obs}(k)$ is the observed, $T_s$ is the sampling time and $E_{PC}(k)$ is the energy dissipated by the PC according to the following equation. If the observed energy becomes negative, the passivity condition has been violated, thus the PC intervenes to guarantee stability according to the following:

\begin{equation}\label{eq:PC}
\begin{aligned}
\mathbf{f}_{cmd}(k) &= \mathbf{\hat{f}}(k) + \alpha (k)\mathbf{v}(k)\\
\alpha(k) &= \left\{\begin{matrix*}[l]
\dfrac{-E_{obs}(k)}{T_{s}\mathbf{v}^{T}(k)\mathbf{v}(k)} & if \ 
\begin{cases}
E_{obs}(k)<0\\
\mathbf{v}^{T}(k)\mathbf{v}(k) > 0\\
\end{cases}\\
0 & otherwise\\
\end{matrix*}\right.\\
E_{PC}(k) &= E_{PC}(k-1) + T_{s}\mathbf{v}^{T}(k) \alpha(k) \mathbf{v}(k)\\
\end{aligned}
\end{equation}

where $\alpha (k)$ is the damping factor and $\mathbf{f}_{cmd}(k)$ is the commanded force to the robot after the dissipation action. For clarity, the explicit time dependence will hereafter be omitted.

\begin{figure}[t]
    \centering
		\includegraphics[width=0.5 \textwidth] {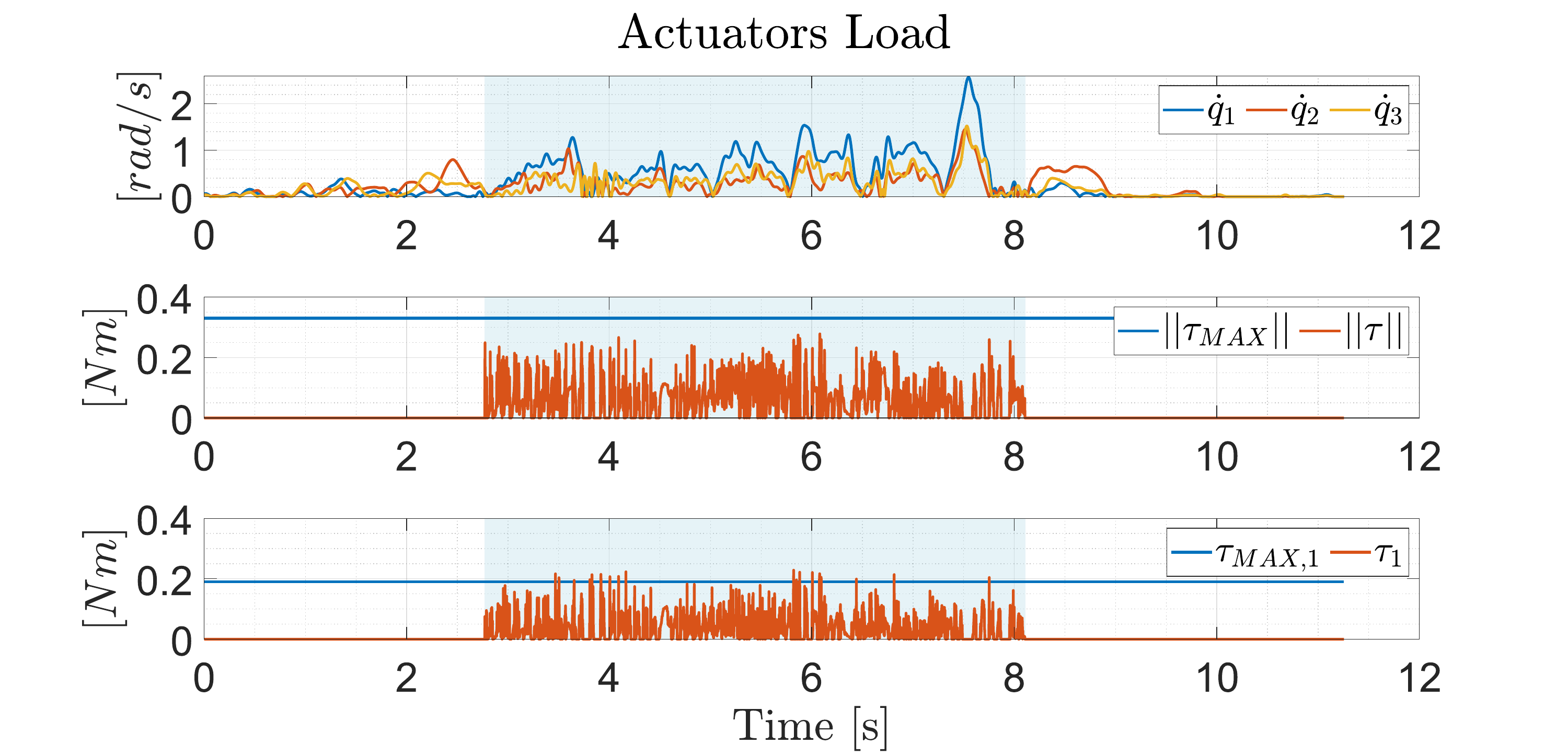}
    \caption{Application of strategy \cite{porcini2023actuator} with anisotropic velocities. On the first row joints' velocity are shown, where $\dot{q}_1$ is always greater than the others during the dissipation. On the second row the limitation in norm of the torques is verified. On the last row the dissipation torque at joint 1 is demonstrated to violate the physical limits.}
    \label{fig:limitazione2023_1}
\end{figure}

Equation \ref{eq:PC} makes explicit that, due to the definition of damping $\alpha(k)$, the PC aims to dissipate the overall observed energy in one sample time. As it was already stated and as it is recognized at the state-of-the-art, this is inconsistent with the system’s physical limitations. A limitation of the damping applied by the PC is necessary to guarantee proper energy dissipation and thus stability.

The most recent and advanced limitation strategy proposes a closed-form solution that takes into account both actuator power-limits and robot configuration \cite{porcini2023actuator}. This limitation relies on a norm-based inequality where the dissipating action is limited by the maximum applicable force, in accordance with actuator power-limits and with the configuration:

\begin{equation}\label{eq:limit_old}
   ||\alpha \mathbf{v}|| \leq ||\mathbf{f}_{MAX}|| \Rightarrow \alpha_{MAX} = \sqrt{\dfrac{{\mathbf{f}}_{MAX}^T {\mathbf{f}}_{MAX} }{\mathbf{v}^T \mathbf{v} }}
\end{equation}

where $\mathbf{f}_{MAX} = J^{-T} \bm{\tau}_{MAX}$ is the maximum feasible force computed from the maximum torques of the actuators thanks to the Jacobian matrix $J$. Note that $\mathbf{f}_{MAX}$ is computed considering a square robot, while for redundant the definition of such a vector requires different computing. It should be noted that $\alpha$ depends on the configuration through the Jacobian (thus it also depends on time) and on the power-limits through $\bm{\tau}_{MAX}$. Thus, each time that $\alpha > \alpha_{MAX}$, $\alpha$ is limited to its maximum value, also limiting the actual dissipated energy according to equation \ref{eq:PC}. The remaining not-yet-dissipated energy is collected for the next sampling steps.

Even if this limitation strategy was shown much more effective than time-based limitations \cite{porcini2023actuator} and more advantageous than model-based \cite{lee2009adjusting,kim2014force} or recursive \cite{ott2011subspace} approaches, it suffers from issues related to the eventual anisotropy in the input velocity vector of the PC (see equation \ref{eq:PC}) or in the maximum foce vector.


\subsection{State-of-the-art Damping Limitation Startegy Issues} \label{sec:limit}

As it was already stated, the damping limitation introduced in \cite{porcini2023actuator} suffers from two main issues. To demonstrate these issues, preliminary experiments were performed that involved contact between a haptic interface and a virtual wall. For more details on the setup, the reader can refer to section \ref{sec:exp}.

\begin{figure}[t]
    \centering
		\includegraphics[width=0.5 \textwidth] {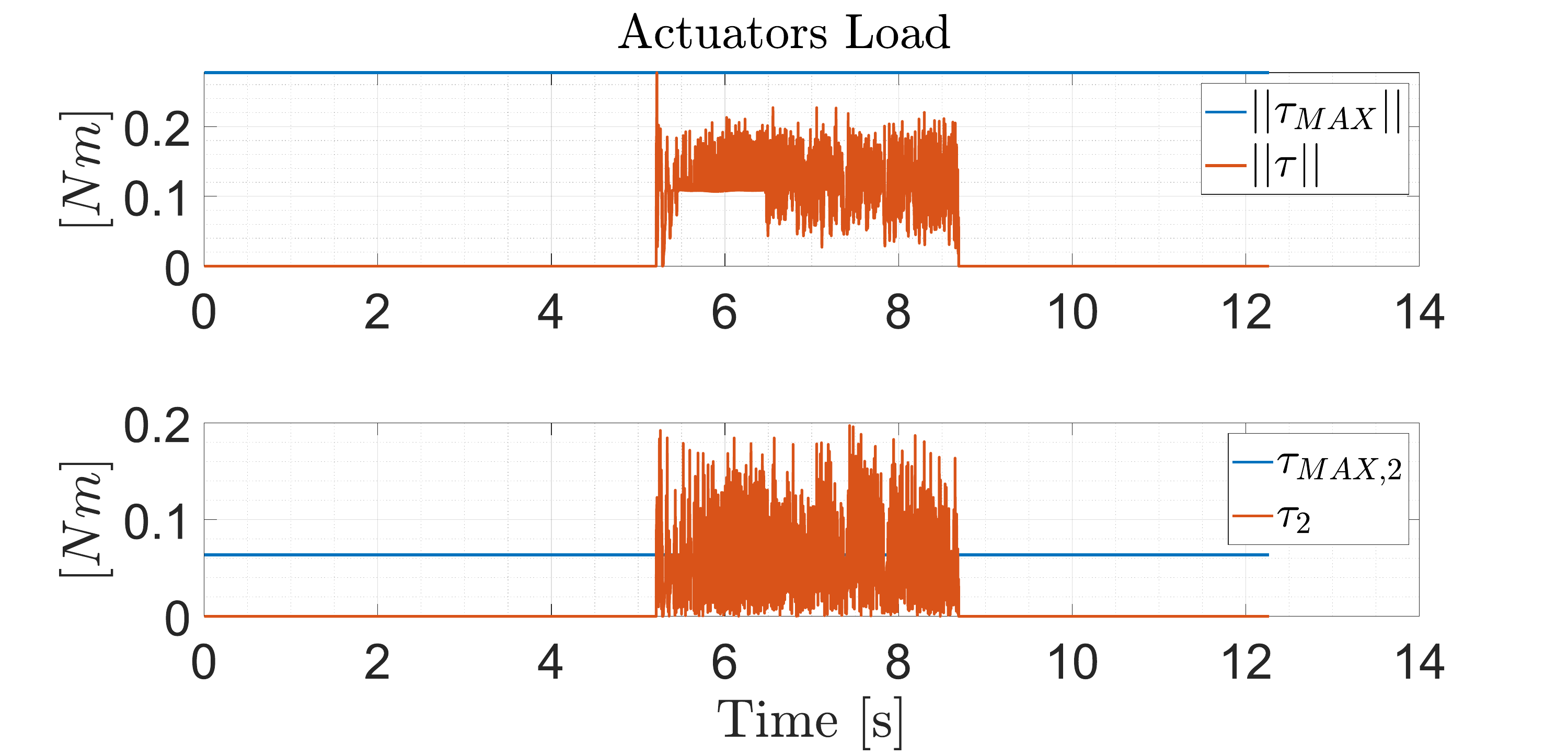}
    \caption{Application of strategy \cite{porcini2023actuator} with anisotropic actuator limits. On the first row, the limitation in the norm of the torques is satisfied. On the second row, the torque at joint 2 – which was constrained to one third of its maximum possibilities – is shown to not satisfy the physical limits.}
    \label{fig:limitazione2023_2}
\end{figure}

The first issue is due to the possible anisotropy of the joint velocity vector. This situation not infrequently occurs when a movement is mainly performed along one direction (e.g. a movement along one axis in the Cartesian space). In such a situation, an anisotropic velocity requires dissipation mainly along the motion direction (see equation \ref{eq:PC}). Nevertheless, even if the norm of the dissipating action is ensured limited by the norm of the maximum force, applying a constraint based on the norm of the power limits vector (which encompasses all directional limits) incorrectly elevates the bound along the true direction of motion. Intuitively, this limitation seems suggesting that moving along a single axis could exploit the power capabilities of the other axes; however, this is clearly incorrect. This behaviour is shown in figure \ref{fig:limitazione2023_1}, where a Cartesian wall-sliding movement was performed such that joint 1 is activated more than joints 2 and 3. As a consequence of equation \ref{eq:limit_old}, the limitation damping strategy is ineffective in limiting dissipation torque over joint 1 to the physical capabilities of the system, while still ensuring norm limitation to be verified.

An analogous failure happens with different joint sizes and different actuators' capabilities. In this case, the vector $\bm{\tau}_{MAX}$ is anisotropic and the norm limitation provided by equation \ref{eq:limit_old} cannot guarantee a one-by-one component limitation. This case is less common for haptic interfaces since they are typically square parallel kinematics robots with symmetric actuators. However, in teleoperation applications, follower manipulators are often robotic arms with different size joints. The effect is analogous to the previous case and it is shown in figure \ref{fig:limitazione2023_2}. In the experiment, joint 2 was constrained with a torque limit equal to one third of the limit of the other two joints. Again, the norm constraint holds, whereas the dissipated torque at joint 2 significantly exceeds its own limit.

\begin{figure}[t]
    \centering
    \includegraphics [width=0.4\textwidth]{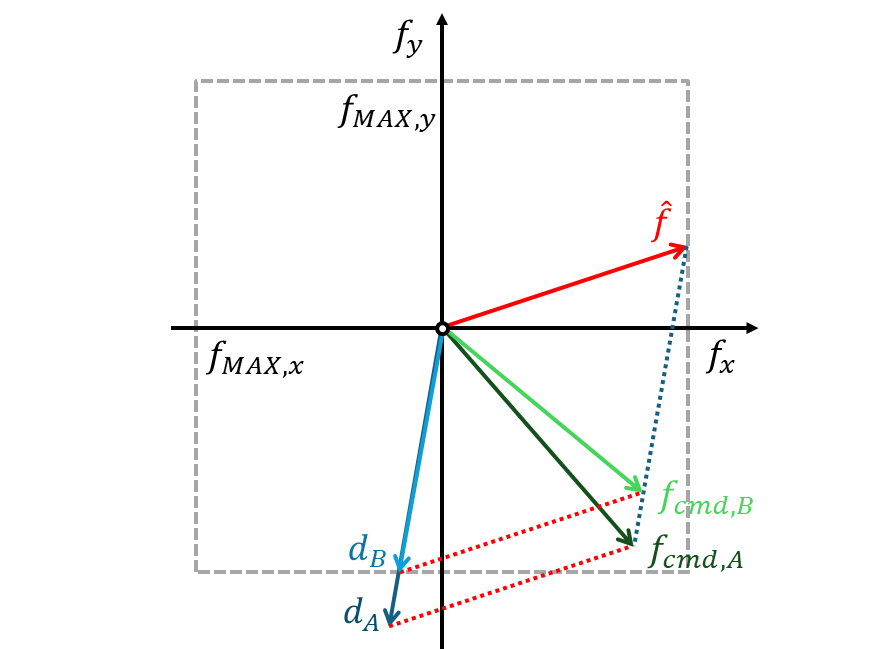}
    \caption{Representation of the ($f_x$,$f_y$) plan. The square highlights the physical capabilities of the system, i.e. references provided to the actuators are realizable if contained in the square. $\mathbf{\hat{f}}$ is the reference force, $\mathbf{d}_A$ is the desired dissipation vector which generates, together with $\mathbf{\hat{f}}$, the commanded force $\mathbf{f}_{cmd,A}$ while $\mathbf{d}_B$ is the limited dissipation vector which generates $\mathbf{f}_{cmd,B}$. Since $\mathbf{f}_{cmd,A}$ is already realizable by the actuators, it should not be necessary to limit the dissipation $\mathbf{d}_A$, but this fact is not taken into account in \cite{porcini2023actuator}.}
    \label{fig:forze1}
\end{figure}

Another important issue of the strategy in \cite{porcini2023actuator} is that the norm limitation is imposed directly on the dissipation vector $\mathbf{d} = \alpha \mathbf{v}$ instead of the sum of the reference $\mathbf{\hat{f}}$ and the dissipation vector. Figure \ref{fig:forze1} clearly shows how this strategy could cause an unnecessary loss of possibilities in dissipation. In fact, assuming $\mathbf{\hat{f}}$ is the force reference and $\mathbf{d}_A$ the desired dissipation vector that exceeds the force limits, the resulting vector $\mathbf{f}_{cmd,A}$ actually commanded to the robot results realizable even if $\mathbf{d}_A$ exceeds the limits. However, according to equation \ref{eq:limit_old} $\mathbf{d}_A$ is limited to $\mathbf{d}_B$, clearly generating an admissible commanded force, but at the cost of over-limiting the dissipating action and causing the need for a further dissipation to guarantee passivity. Thus, the strategy proposed in \cite{porcini2023actuator} is ipso facto excessively conservative.

\section{Priority Direction-based Damping Limitation Strategy}

The dissipation strategy proposed in the present work aims to go beyond the limitations of the state-of-the-art by defining a task priority based dissipation. The optimality criterion adopted to define a priority dissipation direction relies on minimizing the load on the actuators. The strategy attempts to dissipate in this direction as much as possible (according to the newly identified maximum damping). Any additional dissipation requirement is subsequently allocated to the orthogonal hyperplane (without affecting the higher priority direction) up to another newly computed maximum damping. It should be noted that choosing a specific direction is necessary when aiming to impose a consistent damping limit across all axes, valid not only in terms of the norm. Thus, the damping limits are expressed in closed form, eliminating the need for online optimization. These new damping limitations take into account for actuators' power limits across all system axes and maintain robustness against variations in the PC input vector, making them applicable to multi-DoF devices without restrictions due to anisotropy.

\subsection{High Priority Dissipation Direction}

\begin{figure}[t]
    \centering
		\includegraphics[width=0.5 \textwidth] {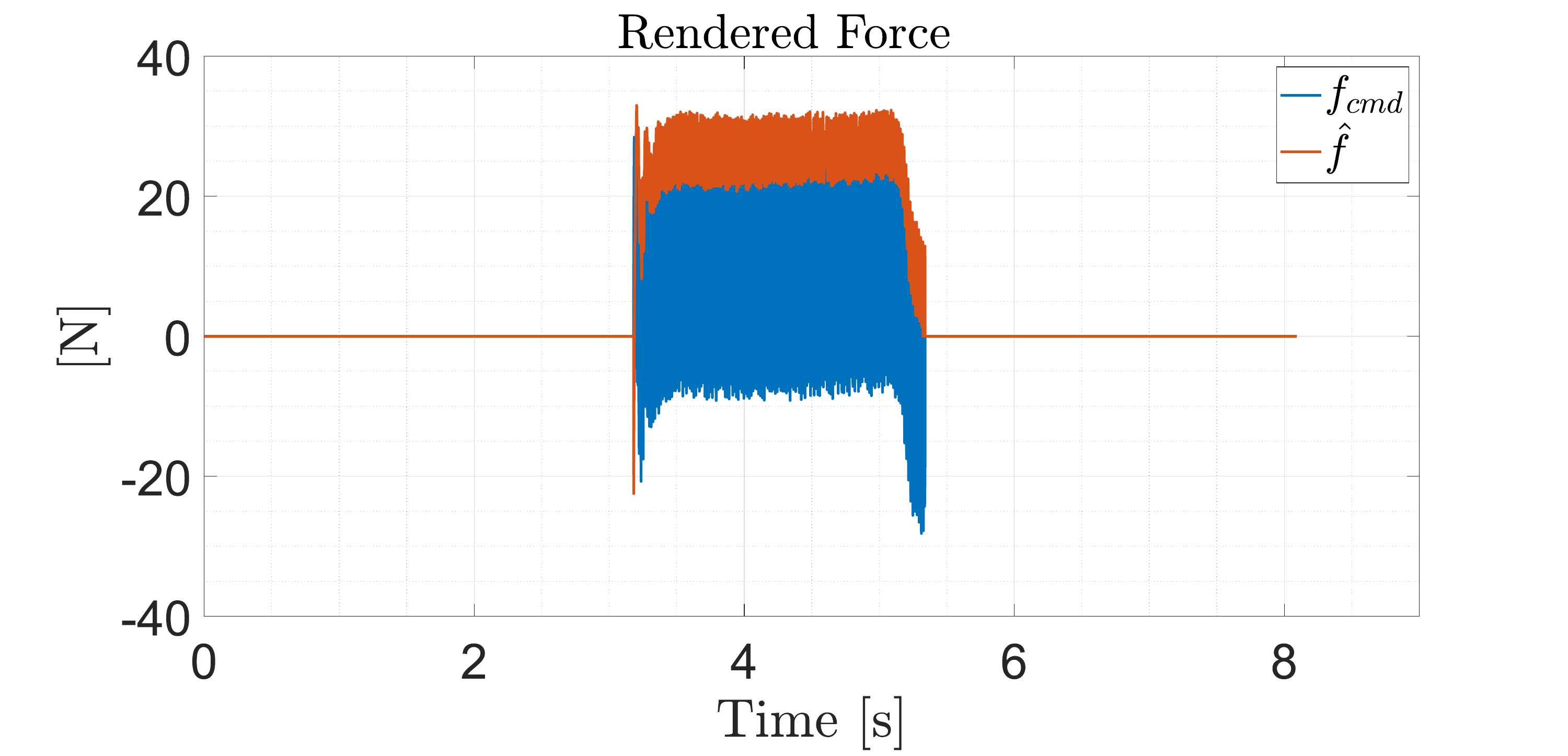}
    \caption{Rendered and dissipated fed back force, respectively $\hat{f}$ and $f_{cmd}$, along the virtual wall axis by using strategy in \cite{preusche2003time}. According to the position of the wall, positive forces are repulsive, while negative are attractive.}
    \label{fig:limite2003}
\end{figure}

The chosen optimal dissipation direction for relieving actuators' load aligns with the rendering direction: when dissipation acts opposite to the rendered vector, it counteracts the reference, thereby lowering the actuators’ required effort.

Defining $\mathbf{\hat{f}}$ as reference due to a haptic interaction, it can be used to define the privileging dissipation direction. It should be noted that $\mathbf{\hat{f}}$ is assumed to be always geometrically limited by the maximum capabilities of the system. In fact, it doesn't make any sense to provide the system with an unfeasible reference. Thus, $\mathbf{\hat{f}}$ can be used to define a projecting matrix $F$ able to project each vector along the direction of interest.

\begin{equation}
    F = \dfrac{\mathbf{\hat{f}} \mathbf{\hat{f}}^T}{\mathbf{\hat{f}}^T \mathbf{\hat{f}}}
\end{equation}

that satisfies projectors' properties:

\begin{equation}
    \begin{aligned}
        & F = F^T \\
        & F^T F = F\\
        & (I - F)^T F = 0
    \end{aligned}
\end{equation}

where $I$ is the identity matrix. This matrix can be used to define the dissipation strategy as follows:

\begin{equation}
\mathbf{f}_{cmd} = \mathbf{\hat{f}} + \alpha F \mathbf{v}
\end{equation}

This direction was already identified by Preusche at al. \cite{preusche2003time} as a transparency-enhancing. However, no limit for the damping was provided with that formulation. Thus, it is not considered that, without limiting the damping factor, it could be generated a dissipation vector greater in norm than the reference vector, giving thus an opposite feedback to the operator. A critical issue arises if the damping factor is left unbounded: the dissipation vector can surpass (if required by an unlimited damping) the reference vector in magnitude, resulting in feedback (sum of the reference and the dissipating action) opposed in direction to the desired rendering. Therefore, in contact with a wall, the operator does not feel a repulsive force \textit{from} the wall, whereas an attractive one \textit{to} the wall. To better understand this concept, Figure \ref{fig:limite2003} shows the plots relative to a preliminary experiment outlined as a simple contact with a virtual wall along one axis. The interaction force $\mathbf{\hat{f}}$ and the passivated commanded force $\mathbf{f}_{cmd}$ are shown. The previously described effect can be easily observed: a high amount of observed energy causes the dissipation vector to surpass the reference, resulting in a sign inversion of the rendered force $\mathbf{f}_{cmd}$. This is clearly an inaccurate representation of the remote environment, which may even exacerbate or feed system instability.

To prevent the PC from generating attractive to the wall references, the damping must be constrained such that the dissipation, aligned with the reference, never exceeds it in magnitude: $||\alpha F \mathbf{v}|| \leq ||\mathbf{\hat{f}}||$. Thus, the dissipation strategy is defined with a new saturation on the damping factor as:

\begin{figure}[t]
    \centering
    \includegraphics [width=0.45\textwidth]{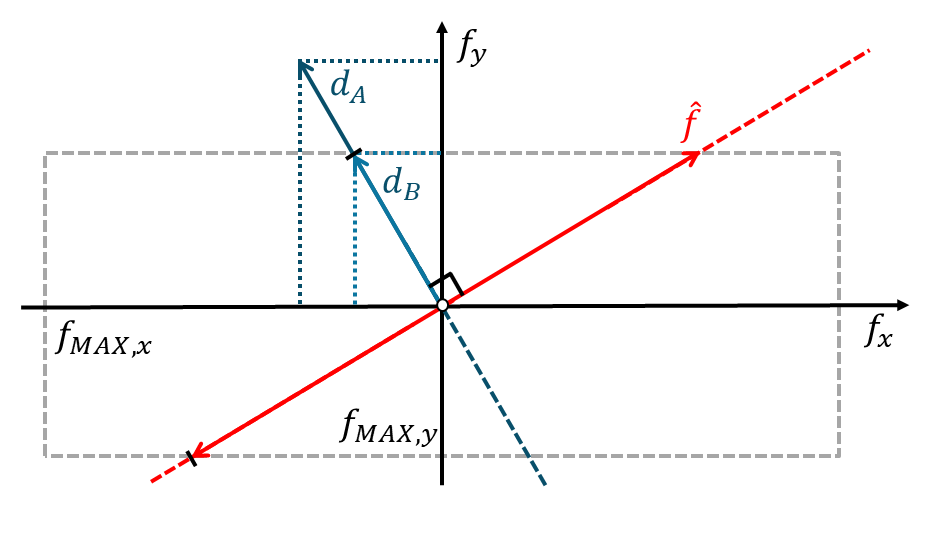}
    \caption{Representation in ($f_x$,$f_x$) plan of the proposed strategy. In red, high-priority dissipation is represented, directed as the reference and limited by the reference magnitude. In blue, the orthogonal low-priority direction is identified. Here, the vector $\mathbf{d}_A$ respects the limit along x-axis, but not along y-axis. Conversely, the proposed limitation for this direction requires $\mathbf{d}_{A}$ to be limited up to $\mathbf{d}_B$, respecting the force constraints.}
    \label{fig:forze2}
\end{figure}

\begin{equation}\label{eq:PC}
\begin{aligned}
\mathbf{f}_{cmd} &= \mathbf{\hat{f}} + \alpha  F \mathbf{v}\\
\alpha &= \left\{\begin{matrix*}[l]
min\left\{\dfrac{-E_{obs}}{T_{s}\mathbf{v}^{T}F\mathbf{v}},\alpha_{MAX}\right\} & if \ 
\begin{cases}
E_{obs}<0\\
\mathbf{v}^{T}F\mathbf{v} > 0\\
\end{cases}\\
0 & otherwise\\
\end{matrix*}\right.\\
&\text{with} \ \alpha_{MAX} = \sqrt{\dfrac{\mathbf{\hat{f}}^T \mathbf{\hat{f}} }{\mathbf{v}^T F \mathbf{v} }}\\
\end{aligned}
\end{equation}

This defines the dissipation with the highest priority, generating a force in the direction of the reference that is less than or equal to it. The effect of this dissipation on the actuators is mainly to lighten their load, decreasing the force that they have to render while maintaining fixed the rendering direction. It is clear that the new passivated force cannot exceed the physical limits of each actuator singularly, thanks to the hypothesis of geometrically limited reference.

If $\alpha > \alpha_{MAX}$ the damping factor is saturated and there still is a residual energy to be dissipated defined as:

\begin{equation}
    E_{res} = E_{obs} + \alpha_{MAX} \mathbf{v}^T F \mathbf{v}
\end{equation}

\subsection{Low Priority Dissipation Direction}

In order to fully exploit the dissipative capabilities of the system and guarantee stability, the residual energy is dissipated with a lower priority strategy. In particular, a dissipation orthogonal to the previous one to avoid interferences is defined using the matrix $(I-F)^T$ to project the actual velocities.

\begin{equation}\label{eq:PC}
\begin{aligned}
\mathbf{f}_{cmd} &= \mathbf{\hat{f}} + \alpha  F \mathbf{v} + \alpha_O (I-F) \mathbf{v}\\
\alpha_O &= \left\{\begin{matrix*}[l]
min\left\{\dfrac{-E_{res}}{T_{s}\mathbf{v}^{T}(I-F)\mathbf{v}},\alpha_{O_{MAX}}\right\} & if \ 
\begin{cases}
E_{res}<0\\
\mathbf{v}^{T}(I-F)\mathbf{v} > 0\\
\end{cases}\\
0 & otherwise\\
\end{matrix*}\right.\\
\end{aligned}
\end{equation}

Even in this case, the damping factor has to be limited to not exceed the maximum capabilities of the actuators. To avoid the same issue as in \cite{porcini2023actuator}, it is necessary to consider the maximum force in each direction separately from the other directions and then verify that the newly introduced dissipation doesn't exceed any. Thus, if $\mathbf{f}_{MAX} = [f_{max_1},...,f_{max_n}]$ is an $n$-dimensional vector, $n$ vectors can be defined such as for each:

\begin{equation}
    \mathbf{f}_{MAX_i} = [0,...0, f_{max_{i}}, 0,...0]^T \ \text{with} \ i = 1,...,n
\end{equation}

It is now possible to define a projection vector along that direction of each maximum for each $i$:

\begin{equation}
    F_{MAX_i} = \dfrac{\mathbf{f}_{MAX_i} \mathbf{f}_{MAX_i}^T}{\mathbf{f}_{MAX_i}^T \mathbf{f}_{MAX_i}}
\end{equation}

Thus, defining $\mathbf{v}_o = (I-F) \mathbf{v}$, the following inequality must hold for each $i$:

\begin{equation}
    ||\alpha_o F_{MAX_i} \mathbf{v}_o|| \le ||\mathbf{f}_{MAX_i}||
\end{equation}

The maximum value for the damping factor is defined as:

\begin{equation}
    \alpha_{o_{MAX}} = 
\min_{i=1,...n} \Bigg\{\dfrac{\mathbf{f}_{MAX_i}^T \mathbf{f}_{MAX_i}}{\mathbf{f}_{MAX_i}^T (I-F) \mathbf{v}}\Bigg\}
\end{equation}

A graphic 2D interpretation is shown in Figure \ref{fig:forze2} as an example. The dissipation with lower priority, along the direction orthogonal to the force reference, generates a vector which allows to dissipate more energy and to continue to stabilize the system.
If even this dissipation is not enough to guarantee stability, i.e. $\alpha_O > \alpha_{O_{MAX}}$, there is an additional residual energy to be dissipated in the next sampling step:

\begin{multline}
    E_{res} = E_{obs} - \alpha_{MAX} \mathbf{v}^T F \mathbf{v} - \alpha_{O_{MAX}} \mathbf{v}^T (I - F) \mathbf{v}
\end{multline}

It should be noted that this formulation is not limited to the impedance-based configuration. On the contrary, as it often occurs for the PC formulation, the admittance-based one is symmetric and can be obtained by just substituting forces with velocities and vice versa.

\section{EXPERIMENTS AND RESULTS} \label{sec:exp}

\begin{figure}[t]
    \centering
    \includegraphics[width=0.49 \textwidth] {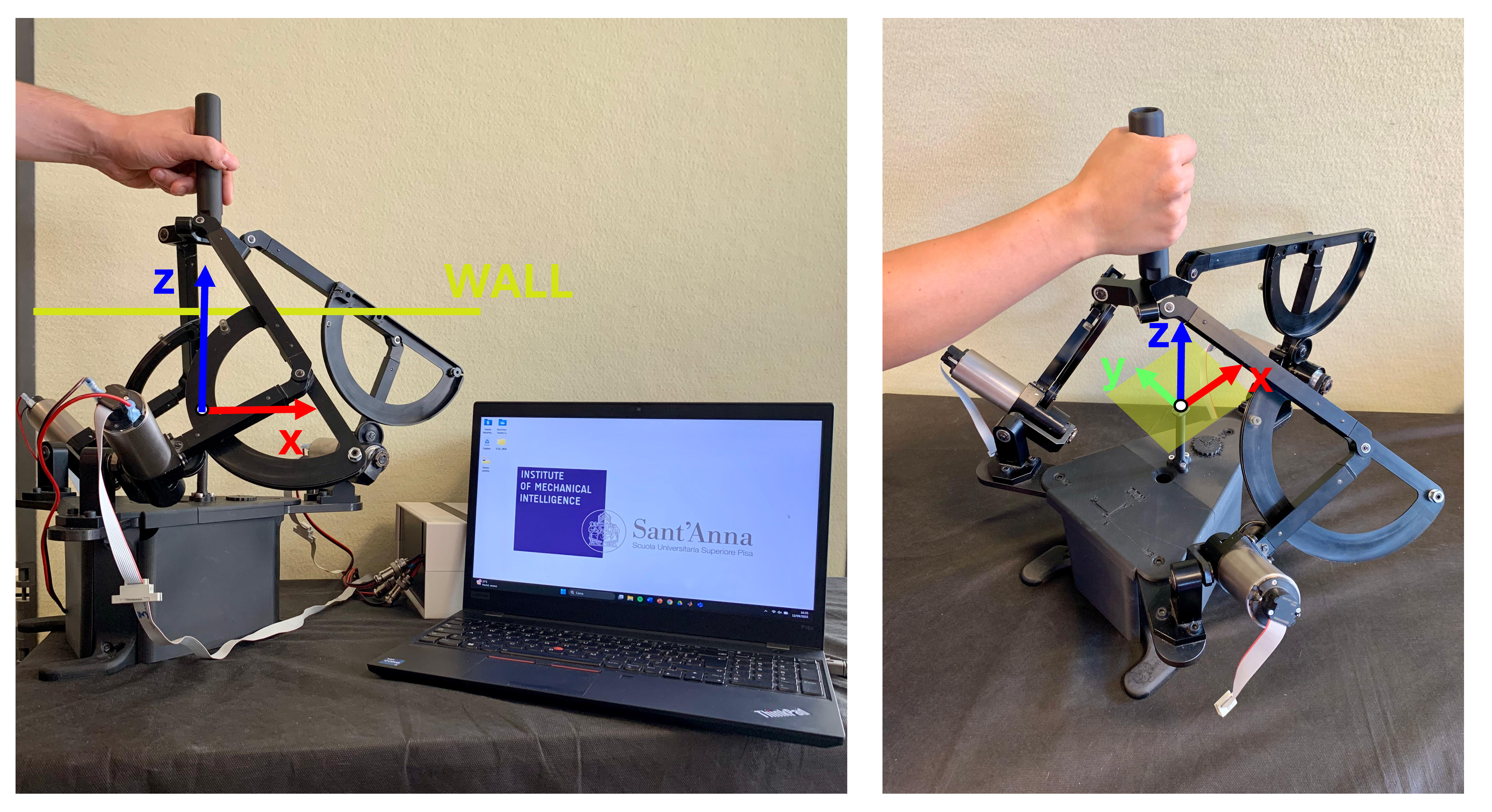}
    \caption{Experimental setup composed by a 3 DoFs parallel haptic interface. The virtual wall is implemented to lay in the xy-plane, parallel to the base, generating a force is in the positive z direction.}
    \label{fig:setup}
\end{figure}

\subsection{Setup}

For all of the experiments proposed in this work, a 3-DoFs custom parallel haptic interface is used \cite{frisoli2004force}. The haptic interface, visible in figure \ref{fig:setup}, has a Delta-like kinematics, in which each leg is a RRR chain and the closest joint to the base is actuated. The motors equipped are FAULHABER DC-Micromotors Series 3890 024 CR. The resulting parallel kinematic chain allows the end-effector to translate in the 3D space keeping fixed the orientation. The control PC is a real-time target machine running \textit{Matlab\textsuperscript{\textregistered} Simulink Real-Time} at a control loop frequency of $5 \ KHz$.

The virtual wall is implemented, with stiffness and damping equal to $K = 20000 N/m$ and $b = 130Ns/m$ respectively, to lie in the xy-plane, parallel to the base of the robot, generating a force along the z direction. In this configuration the wall acts passively and the system is stable.
For the experiments of this section, the wall is forced to behave actively, inverting the sign of the damping factor as in \cite{hannaford2002time}, leading the system to instability.


\begin{figure}[t]
    \centering
		\includegraphics[width=0.49 \textwidth] {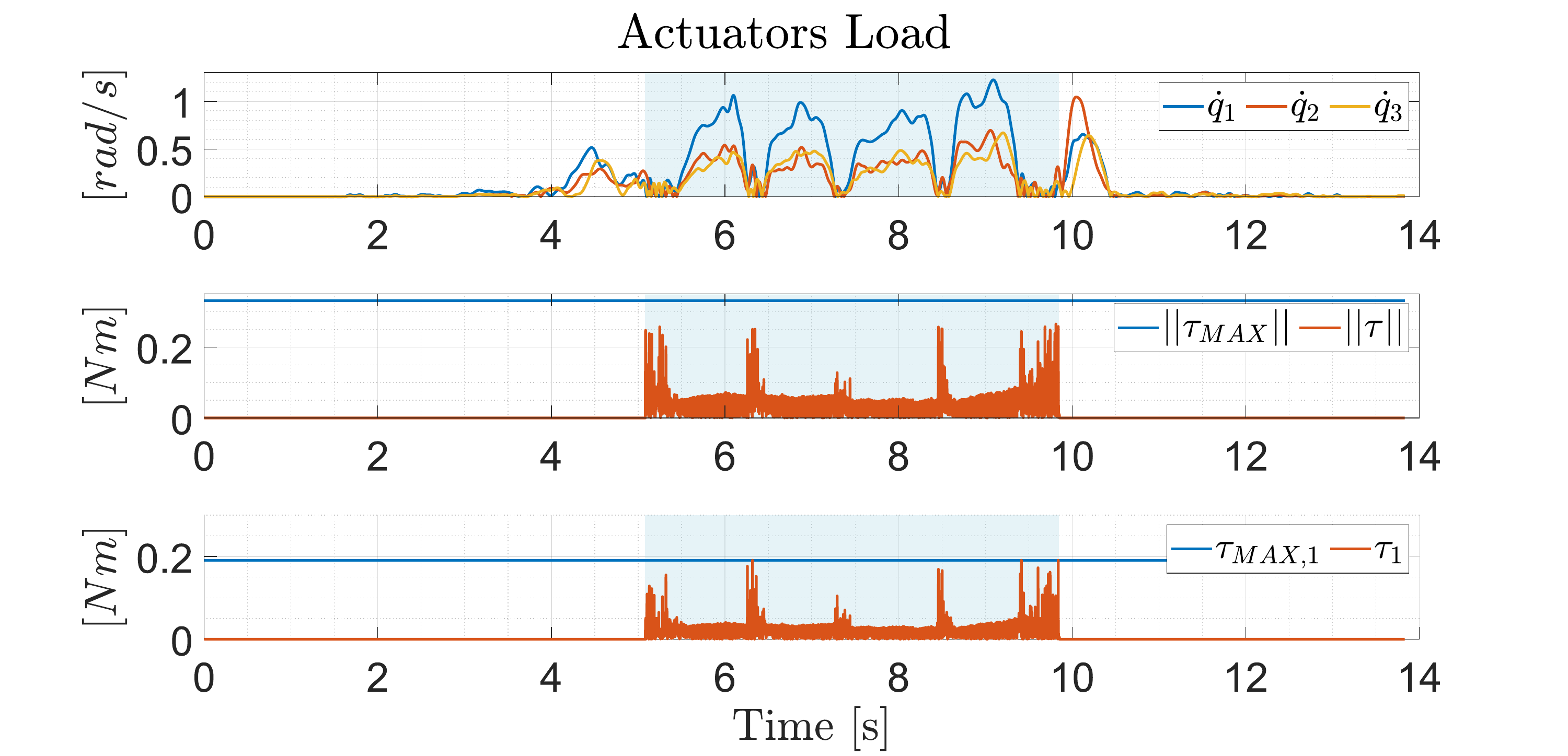}
    \caption{\textit{Experiment 1}: Application of the new strategy with anisotropic velocities. On the first row joints’ velocity are shown, where $\dot{q}_1$ is always greater than the others during the dissipation. On the second row the norm of the actual torques is demonstrated to be lower than the norm of maximum forces. On the last row the torque at joint 1 stays below the practical limits.}
    \label{fig:lim2023_1_new}
\end{figure}

\begin{figure}[t]
    \centering
		\includegraphics[width=0.49 \textwidth] {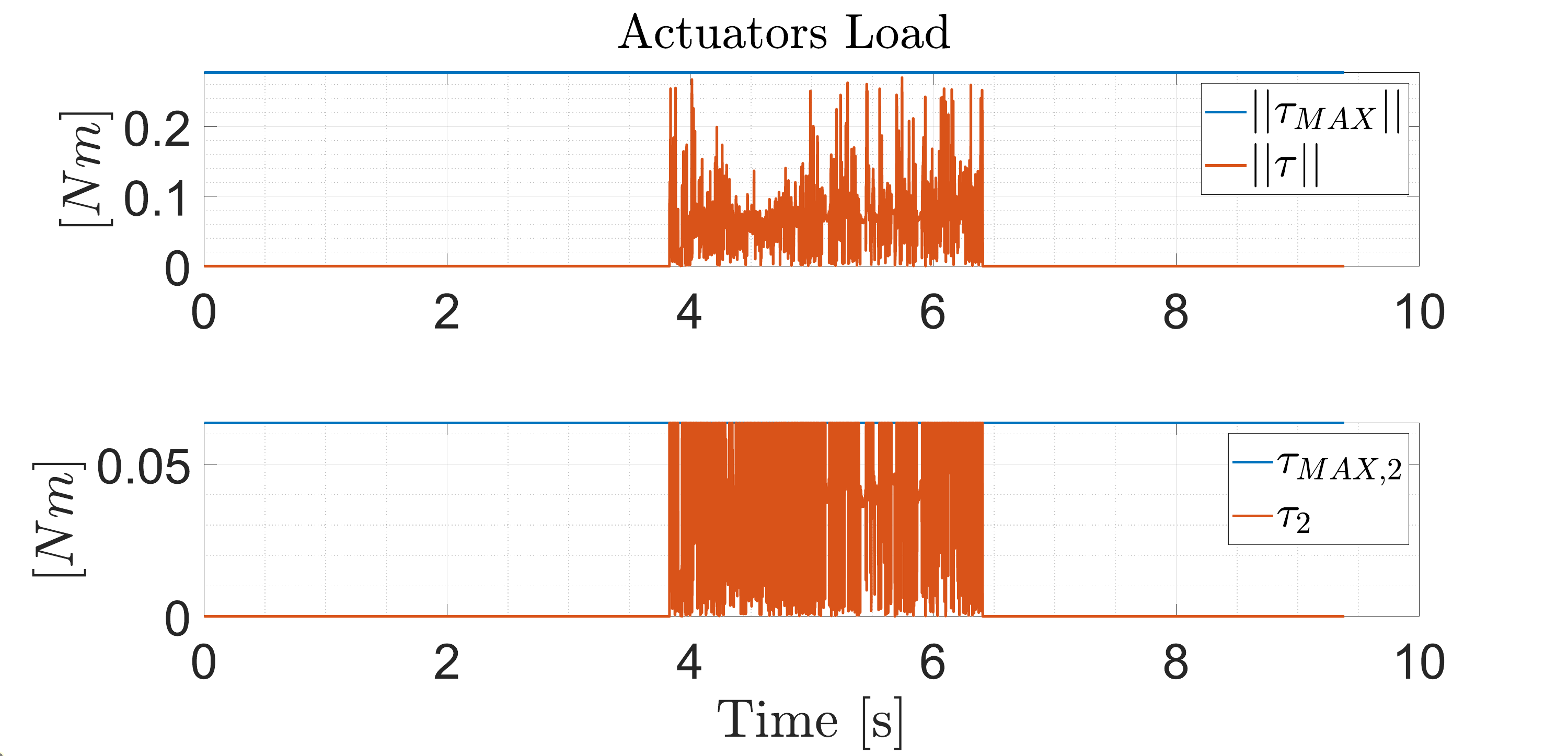}		
    \caption{\textit{Experiment 2}: Application of the new strategy with anisotropic actuator limits. On the first row, the limitation in the norm of the torques is satisfied. On the second row, the torque at joint 2 – which was constrained to one third of its maximum possibilities – is shown to satisfy the physical limits.}
    \label{fig:lim2023_2_new}
\end{figure}

\begin{figure*}[t]
    \centering
				\includegraphics[width=0.49 \textwidth] {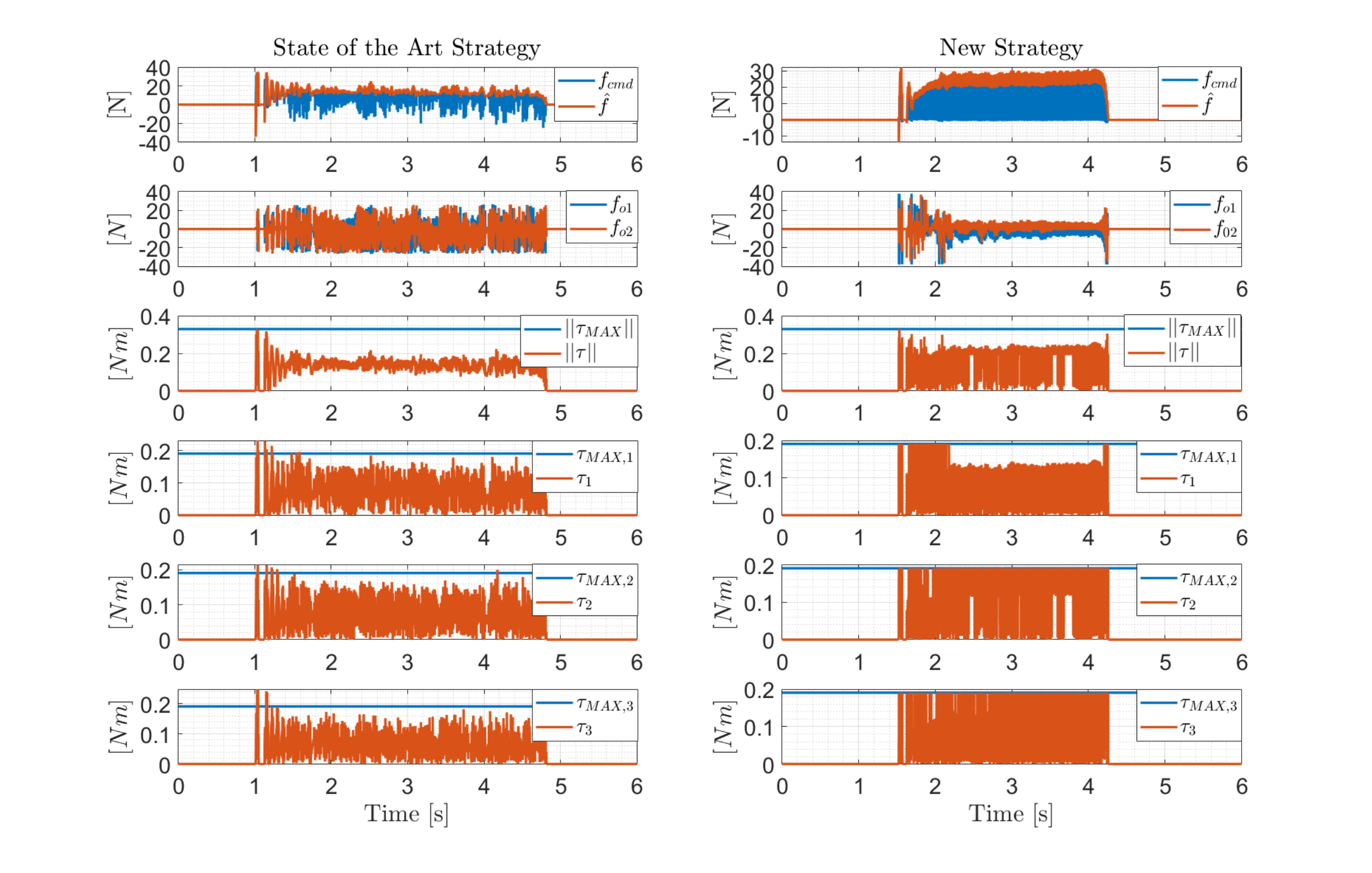}
    \caption{\textit{Experiment 3}: the figure compares the state-of-art strategy with the new one proposed by the authors. On the first row the rendered force from the virtual wall (red) is shown along with the dissipated one (blue). The second row presents the orthogonal dissipation contributions. The remaining rows highlight the actuators load both in norm and for each joint separately (where each maximum was highlighted in blue). }
    \label{fig:comparison}
\end{figure*}

\subsection{Results}

Several experiments were conducted to verify that the limitations of previous methods, identified in Section \ref{sec:limit}, have been overcome. Each experiments consists in establishing a stable contact with the virtual wall for a certain amount of time under different conditions.


\begin{figure}[b!]
    \centering
				\includegraphics[width=0.49 \textwidth] {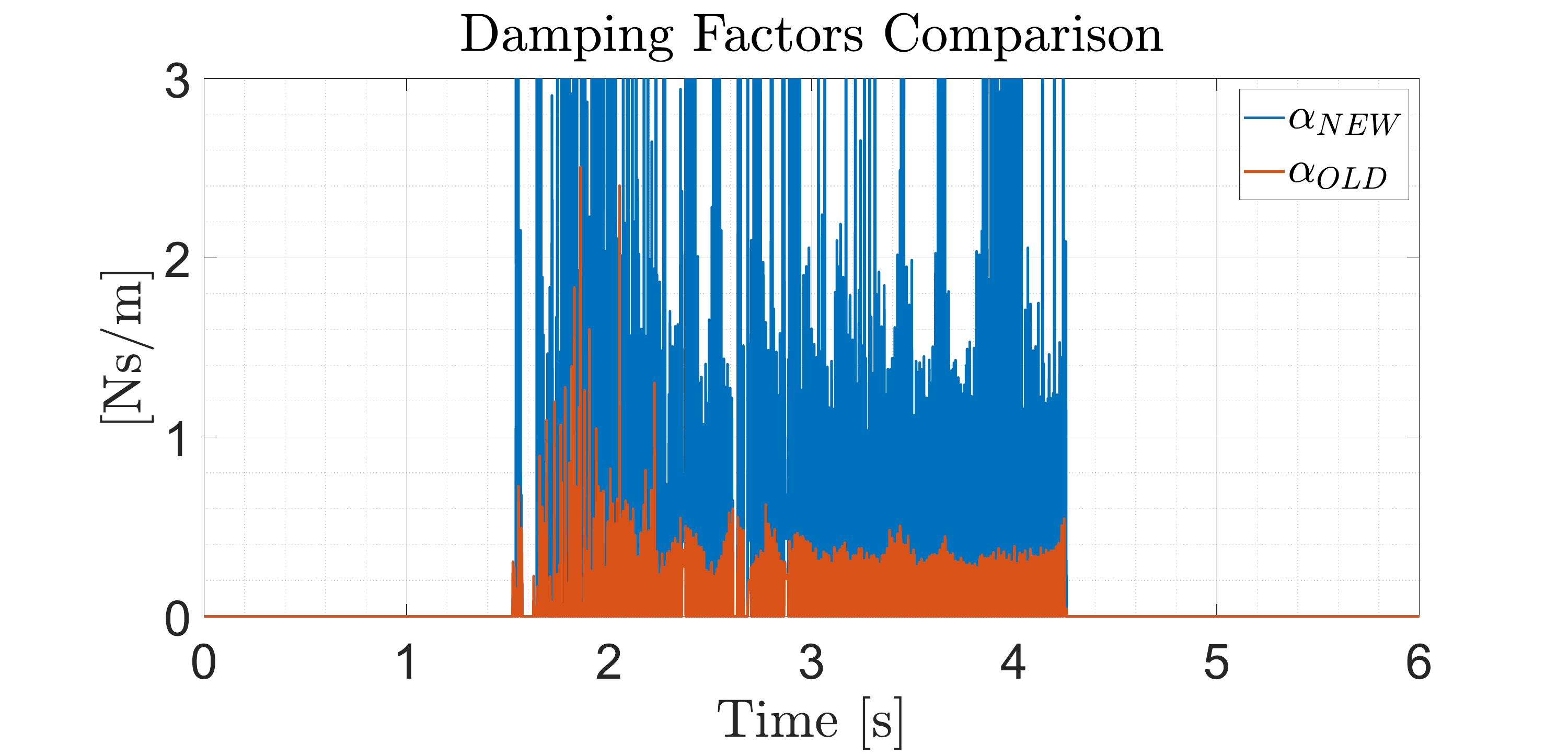}
    \caption{\textit{Experiment 3}: comparison between the damping factors $\alpha$ in the trials with the strategy in \cite{porcini2023actuator} (in red) and with the strategy presented (in blue).}
    \label{fig:aplha}
\end{figure}

Specifically, \textit{Experiments 1 and 2} demonstrate the robustness to the problems identified in Section \ref{sec:limit}. The first experiment is conducted to demonstrate the effectiveness of the limitation in the presence of anisotropic velocities. The task consists of two phases: first, reaching the wall through a movement along the z-axis; then, once in contact, performing a motion on the wall’s xy-plane so as to engage one joint more than the others. On the first row of Figure \ref{fig:lim2023_1_new}, it is shown how the first joint is more active than the others, with higher velocities during all the highlighted in blue phase. This anisotropic behaviour related to the executed movement no longer results in a power-limits constraint violation thanks to the proposed strategy. In fact, in term of torque, each joint is always under the maximum values, element by element and so in norm. In the \textit{Experiments 2}, the condition of actuators with different capabilities is tested. This condition is realized by limiting the second actuator up to one third of its maximum power. As in the previous experiment, figure \ref{fig:lim2023_2_new} shows the second joint (the more limited) respecting the limitation singularly thanks to the new strategy.

Finally, the last experiment, referred to as \textit{Experiment 3}, highlights the differences between the behaviour of the proposed method and the state-of-the-art approach. In particular, the new strategy is now compared with the method proposed in \cite{porcini2023actuator}, which already outperformed other damping limitation strategies. A comparison between the two methods is presented in Figure \ref{fig:comparison}. In this case as well, contact with the wall is performed. The first row shows plots in both conditions of the reference $\mathbf{\hat{f}}$ and the passivated rendering $\mathbf{f}_{cmd}$ forces. It is easy to notice that the contact achieved with the state-of-the-art strategy is more difficult to maintain and is characterized by slight oscillations, with an average contact force value of approximately 17 N. In contrast, the proposed strategy allows a nearly constant contact, after the initial stabilization phase, with an average contact force of 25 N. Also, for the state-of-the-art strategy, it is visible the inversion of sign in the passivated rendering force that pushes into the wall. This is presumably the reason besides the oscillations. Conversely, the proposed approach avoids this behaviour.

On the second row of Figure \ref{fig:comparison}, the orthogonal components to the high-priority dissipation are visible for both strategies. Even if the order of magnitude is similar, it is evident the prioritized dissipation in the proposed approach. In particular, the orthogonal components are high when entering and exiting the walls, i.e. the moments in which dissipation is required, but the rendering force is low due to the position with respect the wall. Conversely, the state-of-the-art approach balances the dissipation along all the axes since no direction constraints were introduced.

The remaining rows of the figure show the joint torques both in norm and singularly, again illustrating that the state-of-the-art strategy excessively stresses each joint, which does not occur with the new method. It should be noted that both approaches guarantee norm constraint to be verified, but only the proposed one is able to ensure to not overcome the power-limits of the actuators singularly.

Finally, figure \ref{fig:aplha} compares the overall damping applied by the two strategies. The damping value calculated according to the proposed strategy results always higher than the damping calculated using the state-of-the-art strategy, demonstrating a greater utilization of the system’s dissipative capacity through the proposed method and reducing the unnecessary conservatism introduced by limiting directly the dissipating action. Furthermore, by also involving the lower-priority strategy, the new approach is able to further exploit the system’s dissipative capacity, ensuring passivity at every sampling instant.

\section{DISCUSSION AND CONCLUSIONS} \label{sec:disc}

In conclusion, this work has introduced a dissipation strategy that exploits newly defined damping limitations to ensure consistent stabilizing behaviour of the PC. Specifically, this strategy is able to manage anisotropic conditions arising from either the robot's motion or its actuators' capabilities. This closed-form solution also avoids environment-attractive reference generation and reduces the conservatism introduced by other state-of-the-art damping limitation strategies. The importance of prioritizing dissipation along an optimal direction, corresponding to the reference direction, in order to minimize actuator load, has also been emphasized.

The analysis of the state of the art has shown that the solutions proposed so far to respect the physical limits of the actuators do not provide satisfactory performance in the presence of anisotropy, either due to the robot’s motion or to differences in joint sizes. To overcome these limitations, a solution has been proposed defining a two-level prioritized dissipation: the first priority along the reference direction and the second along its orthogonal hyperplane.

The new strategy was tested on a parallel haptic interface interacting with a virtual wall exhibiting active and destabilizing behavior. Experiments demonstrated that the proposed solution ensures a more realistic reproduction of the remote interaction, following the task-priority dissipation strategy and optimizing actuator load, while at the same time limiting damping to respect joint-by-joint power constraints even in the presence of anisotropy.

Future work will focus on applying the method with state-of-the-art observers \cite{dinc2024relaxing} in teleoperation systems where actuator power limits are still not fully considered \cite{panzirsch2024enhancing}.

\section*{Acknowledgement}

This research was produced with the funding of the European Union - Next Generation EU, based on the public notice for the presentation of intervention proposals for the creation and strengthening of ``Innovation ecosystems", construction of ``Territorial R\&D leaders", within the project PNRR MUR - Mission 4 - Component 2 - Investment 1.5, ``THE - Tuscany Health Ecosystem (code ECS00000017) - Spoke 9".

\addtolength{\textheight}{-0cm}   










\bibliographystyle{IEEEtran}
\bibliography{references}

\vspace{-4cm}

\end{document}